\documentclass[11pt,a4paper]{article}
\usepackage[hyphenbreaks]{breakurl}
\usepackage[hyperref]{emnlp2020}
\usepackage{times}
\usepackage{latexsym}
\usepackage{dingbat}
\usepackage{booktabs}
\usepackage{multirow}

\usepackage{microtype}

\aclfinalcopy

\title{Which *BERT? A Survey Organizing Contextualized Encoders}

\author{Patrick Xia ~~~~~ Shijie Wu ~~~~~ Benjamin Van Durme\\
     Johns Hopkins University \\
  \texttt{\{paxia, vandurme\}@cs.jhu.edu, shijie.wu@jhu.edu}}

\date{}

\begin{document}
\maketitle
\begin{abstract}

Pretrained contextualized text encoders are now a staple of the NLP community.
We present a survey on language representation learning with the aim of
consolidating a series of shared lessons learned across a variety of 
recent efforts. While significant advancements continue at a rapid 
pace, we find that enough has now been discovered, in different 
directions, that we can begin to organize advances according to common 
themes. Through this organization, we highlight important considerations
when interpreting recent contributions and choosing which model to use.

\end{abstract}

\section{Introduction}

A couple years ago, \citet[ELMo]{peters-etal-2018-deep} won the NAACL Best Paper Award for creating strong performing, task-agnostic sentence representations due to large scale unsupervised pretraining. Days later, its high level of performance was surpassed by \citet{radford18GPT} which boasted representations beyond a single sentence and finetuning flexibility. 
This instability and competition between models has been a recurring theme for researchers and practitioners who have watched the rapidly narrowing gap between text representations and language understanding benchmarks. However, it has not discouraged research. Given the recent flurry of models, we often ask: ``\textbf{What, besides state-of-the-art, does this newest paper contribute? Which encoder should \textit{we} use?}''

The goals of this survey are to outline the areas of progress, relate contributions in text encoders to ideas from other fields, describe how each area is evaluated, and present considerations for practitioners and researchers when choosing an encoder. 
This survey does not intend to compare specific model metrics, as tables from other works provide comprehensive insight. For example, Table 16 in \citet{raffel2019exploring} compares the scores on a large suite of tasks of different model architectures, training objectives, and hyperparameters, and  Table 1 in \citet{rogers2020primer} details early efforts in model compression and distillation. We also recommend other closely related surveys on contextualized word representations \cite{Smith2019ContextualWR, rogers2020primer, liu2020survey}, transfer learning in NLP \cite{ruder-etal-2019-transfer}, and integrating encoders into NLP applications \cite{Wolf2019HuggingFacesTS}. Complementing these existing bodies of work, we look at the ideas and progress in the scientific discourse for text representations from the perspective of discerning their differences.

We organize this paper as follows. 
\S\ref{sec:background} provides brief background on encoding, training, and evaluating text representations. 
\S\ref{sec:pretraining} identifies and analyzes two classes of pretraining objectives. 
In \S\ref{sec:efficiency}, we explore faster and smaller models and architectures in both training and inference. 
\S\ref{sec:data} notes the impact of both quality and quantity of pretraining data. 
\S\ref{sec:interpretability} briefly discusses efforts on probing encoders and representations with respect to linguistic knowledge. 
\S\ref{sec:multilinguality} describes the efforts into training and evaluating multilingual representations. 
Within each area, we conclude with high-level observations and discuss the evaluations that are used and their shortcomings. 

We conclude in \S\ref{sec:discussion} by making recommendations to researchers: publicizing negative results in this area is especially important owing to the sheer cost of experimentation and to ensure evaluation reproducibility. In addition, probing studies need to focus not only on the models and tasks, but also on the pretraining data. We pose questions for users of contextualized encoders, like whether the compute requirement of a model is worth the benefits. We hope our survey serves as a guide for both NLP researchers and practitioners, orienting them to the current state of the field of contextualized encoders and differences between models.

\section{Background}
\label{sec:background}

\paragraph{Encoders} Pretrained \textit{text encoders} take as input a sequence of tokenized\footnote{Unlike traditional word-level tokenization, most works decompose text into \textit{subtokens} from a fixed vocabulary using some variation of byte pair encoding \cite{Gage:1994:NAD:177910.177914, schuster12wordpiece, sennrich-etal-2016-neural}} text, which is encoded by a multi-layered neural model. The representation of each (sub)token, $x_t$, is either the set of hidden weights, $\{h^{(l)}_t\}$ for each layer $l$, or its weight on just the top layer, $h^{(-1)}_t$. Unlike fixed-sized word, sentence, or paragraph representations, the produced \textit{contextualized representations} of the text depends on the length of the input text. Most encoders use the Transformer architecture \cite{Vaswani17Transformer}.

\paragraph{Transfer: The Pretrain-Finetune Framework}
While text representations can be learned in any manner, ultimately, they are evaluated using specific \textit{target tasks}. Historically, the learned representations (e.g. word vectors) were used as initialization for task-specific models. \citet{Dai15Sem} are credited with using \textit{pretrained} language model outputs as initialization, \citet{mccann2017learned} use pretrained outputs from translation as frozen word embeddings, and \citet{howard-ruder-2018-universal} and \citet{radford18GPT} demonstrate the effectiveness of \textit{finetuning} to different target tasks by updating the full (pretrained) model for each task. We refer to the embeddings produced by the pretrained models (or encoders) as contextualized text representations. As our goal is to discuss the encoders and their representations, we do not cover the innovations in finetuning \cite[\textit{inter alia}]{liu-etal-2015-representation, ruder-etal-2019-transfer, Phang2018SentenceEO,liu-etal-2019-multi, Zhu2020FreeLB:}. 

\paragraph{Evaluation} Widely adopted evaluations of text representations relate them to downstream natural language understanding (NLU) benchmarks. This full-stack process necessarily conflates representation power with finetuning strategies. Common language understanding benchmarks include (1) a diverse suite of sentence-level tasks covering paraphrasing, natural language inference, sentiment, and linguistic acceptability (GLUE) and its more challenging counterpart with additional commonsense and linguistic reasoning tasks (SuperGLUE) \cite{wang2018glue,NIPS2019_8589, clark-etal-2019-boolq, de_Marneffe_Simons_Tonhauser_2019, roemmele2011choice, khashabi-etal-2018-looking, zhang2018record, dagan2006pascal, bar2006second, giampiccolo2007third, bentivogli2009fifth, pilehvar-camacho-collados-2019-wic, rudinger-etal-2018-gender,poliak-etal-2018-collecting-diverse, levesque2011winograd}; (2) crowdsourced questions derived from Wikipedia articles \cite[SQuAD]{rajpurkar-etal-2016-squad, rajpurkar-etal-2018-know}; and (3) multiple-choice reading comprehension \cite[RACE]{lai-etal-2017-race}.

\section{Area I: Pretraining Tasks}
\label{sec:pretraining}
To utilize data at scale, pretraining tasks are typically self-supervised. We categorize the contributions into two types: \textit{token prediction} (over a large vocabulary space) and \textit{nontoken prediction} (over a handful of labels). In this section, we discuss several empirical observations. While token prediction is clearly important, less clear is which variation of the token prediction task is the best (or whether it even matters). Nontoken prediction tasks appear to offer orthogonal contributions that marginally improve the language representations. We emphasize that in this section, we seek to outline the primary efforts in pretraining objectives and not to provide a comparison on a set of benchmarks.\footnote{See \citet{raffel2019exploring} for comprehensive experiments.}

\subsection{Token Prediction}

Predicting (or generating) the next word has historically been equivalent to the task of language modeling. Large language models perform impressively on a variety of language understanding tasks while maintaining their generative capabilities \cite{radford18GPT, radford2019language, Keskar2019CTRLAC, brown2020language}, often outperforming contemporaneous models that use additional training objectives.

ELMo \cite{peters-etal-2018-deep} is a BiLSTM model with a language modeling objective for the next (or previous) token given the forward (or backward) history. This idea of looking at the full context was further refined as a cloze\footnote{A cloze task is a fill-in-the-blank task.} task \cite{baevski-etal-2019-cloze}, or as a denoising Masked Language Modeling (MLM) objective \cite[BERT]{devlin-etal-2019-bert}. MLM replaces some tokens with a \texttt{[mask]} symbol and provides both right and left contexts (bidirectional context) for predicting the masked tokens. The bidirectionality is key to outperforming a unidirectional language model on a large suite of natural language understanding benchmarks \cite{devlin-etal-2019-bert, raffel2019exploring}.

The MLM objective is far from perfect, as the use of \texttt{[mask]} introduces a pretrain/finetune vocabulary discrepancy. \citet{devlin-etal-2019-bert} look to mitigate this issue by occasionally replacing \texttt{[mask]} with the original token or sampling from the vocabulary. \citet{Yang19XLNet} convert the discriminative objective into an autoregressive one, which allows the \texttt{[mask]} token to be discarded entirely. Naively, this would result in unidirectional context. By sampling permutations of the factorization order of the joint probability of the sequence, they preserve bidirectional context. Similar ideas for permutation language modeling (PLM) have also been studied for sequence generation \cite{Stern19InsertTrans, Chan19KERMIT, Gu2019Insertion}. The MLM and PLM objectives have since been unified architecturally \cite{song2020mpnet, bao2020unilmv2} and mathematically \cite{Kong2020A}.

ELECTRA \cite{Clark2020ELECTRA:} replaces \texttt{[mask]} through the use of a small generator (trained with MLM) to sample a real token from the vocabulary. The main encoder, a discriminator, then determines whether each token was replaced. 

A natural extension would mask units that are more linguistically meaningful, such as rarer words,\footnote{\citet{Clark2020ELECTRA:} report negative results for rarer words.} whole words, or named entities \cite{devlin-etal-2019-bert, Sun2019ERNIEER}. This idea can be simplified to \textit{random} spans of texts \cite{Yang19XLNet, Song2019MASSMS}. Specifically, \citet{joshi2019spanbert} add a reconstruction objective which predicts the masked tokens using only the span boundaries. They find that masking random spans is more effective than masking linguistic units. 

An alternative architecture uses an encoder-decoder framework (or denoising autoencoder) where the input is a corrupted (masked) sequence the output is the full original sequence \cite{wang-etal-2019-denoising, lewis-etal-2020-bart, raffel2019exploring}. 

\subsection{Nontoken Prediction}

 \citet{bender-koller-2020-climbing} argue that for the goal of natural language understanding, we cannot rely purely on a language modeling objective; there must be some grounding or external information that relates the text to each other or to the world. One solution is to introduce a secondary objective to directly learn these biases. 

Self-supervised discourse structure objectives, such as text order, has garnered significant attention. To capture relationships between two \textit{sentences},\footnote{\textit{Sentence} unfortunately refers to a text segment containing no more than a fixed number of subtokens. It may contain any (fractional) number of real sentences.} \citet{devlin-etal-2019-bert} introduce the next sentence prediction (NSP) objective. In this task, either sentence \texttt{B} follows sentence \texttt{A} or \texttt{B} is a random negative sample. Subsequent works showed that this was not effective, suggesting the model simply learned topic \cite{Yang19XLNet, Liu19RoBERTa}.  \citet{Jernite2017DiscourseBasedOF} propose a sentence order task of predicting whether \texttt{A} is before, after, or unrelated to \texttt{B}, and \citet{Wang2020StructBERT:} and \citet{Lan2020ALBERT:} use it for pretraining encoders. They report that (1) understanding text order does contribute to improved language understanding; and (2) harder-to-learn pretraining objectives are more powerful, as both modified tasks have lower intrinsic performance than NSP. It is still unclear, however, if this is the best way to incorporate discourse structure, especially since these works do not use real sentences.

Additional work has focused on effectively incorporating multiple pretraining objectives. \citet{sun2019ernie} use multi-task learning with \textit{continual pretraining} \cite{hashimoto-etal-2017-joint}, which incrementally introduces newer tasks into the set of pretraining tasks from word to sentence to document level tasks. Encoders using visual features (and evaluated only on visual tasks) jointly optimize multiple different masking objectives over both token sequences and regions of interests in the image \cite{tan-bansal-2019-lxmert}.\footnote{Table 5 in \citet{Su2020VL-BERT:} provides a recent summary of efforts in visual-linguistic representations.} 

Prior to token prediction, discourse information has been used in training sentence representations. \citet{conneau-EtAl:2017:EMNLP2017, conneau-etal-2018-cram} use natural language inference sentence pairs, \citet{Jernite2017DiscourseBasedOF} use discourse-based objectives of sentence order, conjunction classifier, and next sentence selection, and \citet{nie-etal-2019-dissent} use discourse markers. While there is weak evidence suggesting that these types of objectives are less effective than language modeling \cite{wang-etal-2019-tell}, we lack fair studies comparing the relative influence between the two categories of objectives. 

\subsection{Comments on Evaluation}

We reviewed the progress on pretraining tasks, finding that token prediction is powerful but can be improved further by other objectives. Currently, successful techniques like span masking or arbitrarily sized ``sentences'' are linguistically unmotivated. We anticipate future work to further incorporate more meaningful linguistic biases in pretraining.

Our observations are informed by evaluations that are compared across different works. \textbf{These benchmarks on downstream tasks do not account for ensembling or finetuning and can only serve as an approximation for the differences between the models.} For example, \citet{jiang-etal-2020-smart} develop a finetuning method over a supposedly weaker model which leads to gains in GLUE score over reportedly stronger models. Furthermore, these evaluations aggregate vastly different tasks. Those interested in the best performance should first carefully investigate metrics on their specific task. Even if models are finetuned on an older encoder,\footnote{This the case with retrieval-based QA \cite{guu2020realm, herzig-etal-2020-tapas}, which builds on BERT.} it may be more cost-efficient and enable fairer future comparisons to reuse those over restarting the finetuning or reintegrating new encoders into existing models when doing so does not necessarily guarantee improved performance. 

\section{Area II: Efficiency}
\label{sec:efficiency}

As models perform better but cost more to train, some have called for research into efficient models to improve deployability, accessibility, and reproducibility \cite{amodei18compute, strubell-etal-2019-energy, Schwartz2019GreenA}.  Encoders tend to scale effectively \cite{Lan2020ALBERT:, raffel2019exploring, brown2020language}, so efficient models will also result in improvements over inefficient ones of the same size. In this section, we give an overview of several efforts aimed to decrease the computation budget (time and memory usage) during training and inference of text encoders. While these two axes are correlated, reductions in one axis do not always lead to reductions in the other. 

\subsection{Training}

One area of research decreases wall-clock training time through more compute and larger batches. \citet{You2020Large} reduce the time of training BERT by introducing the LAMB optimizer, a large batch stochastic optimization method adjusted for attention models. \citet{rajbhandari2020zero} analyze memory usage in the optimizer to enable parallelization of models resulting in higher throughput in training. By reducing the training time, models can be practically trained for longer, which has also been shown to lead to benefits in task performance \cite[\textit{inter alia}]{Liu19RoBERTa, Lan2020ALBERT:}.

Another line of research reduces the compute through attention sparsification (discussed in \S\ref{sec:efficiency:space}) or increasing the convergence rate \cite{Clark2020ELECTRA:}. These works report hardware and estimate the reduction in floating point operations (FPOs).\footnote{We borrow this terminology from \citet{Schwartz2019GreenA}.} 
These kinds of speedup are orthogonal to hardware parallelization and are most encouraging as they pave the path for future work in \textit{efficient} training.

Note that these approaches do not necessarily affect the latency to process a single example nor the compute required during inference, which is a function of the size of the computation graph.

\subsection{Inference}
\label{sec:efficiency:space}

Reducing model size without impacting performance is motivated by lower inference latency, hardware memory constraints, and the promise that naively scaling up dimensions of the model will improve performance. Size reduction techniques produce smaller and faster models, while occasionally improving performance. \citet{rogers2020primer} survey BERT-like models and present in Table 1 the differences in sizes and performance across several models focused on inference efficiency.

Architectural changes have been explored as one avenue for reducing either the model size or inference time. In Transformers, the self-attention pattern scales quadratically in sequence length. To reduce the asymptotic complexity, the self-attention can be sparsified: each token only attending to a small ``local'' set \cite{Vaswani17Transformer, child2019sparsetransformer, sukhbaatar-etal-2019-adaptive}. This has further been applied to pretraining on longer sequences, resulting in sparse contextualized encoders \cite[\textit{inter alia}]{qiu2019blockwise, ye2019bptransformer, Kitaev2020Reformer:, Beltagy2020Longformer}. Efficient Transformers is an emerging subfield with applications beyond NLP; \citet{tay2020efficient} survey 17 Transformers that have implications on efficiency.

Another class of approaches carefully selects weights to reduce model size. \citet{Lan2020ALBERT:} use low-rank factorization to reduce the size of the embedding matrices, while \citet{Wang2019StructuredPO} factorize other weight matrices. Additionally, parameters can be shared between layers  \cite{dehghani2018universal, Lan2020ALBERT:} or between an encoder and decoder \cite{raffel2019exploring}. However, models that employ these methods do not always have \textit{smaller} computation graphs. This greatly reduces the usefulness of parameter sharing compared to other methods that additionally offer greater speedups relative to the reduction in model size. 

Closely related, model pruning \cite{Denil2013PredictingPI, Han2015LearningBW, Frankle2018TheLT} during training or inference has exploited the overparameterization of neural networks by removing up to 90\%-95\% parameters. This approach has been successful in not only reducing the number of parameters, but also improving performance on downstream tasks. Related to efforts for pruning deep networks in computer vision \cite{Huang2016DeepNW}, layer selection and dropout during both training and inference have been studied in both LSTM \cite{liu-etal-2018-efficient-contextualized} and Transformer \cite{Fan2020Reducing} based encoders. These also have a regularization effect resulting in more stable training and improved performance. There are additional novel pruning methods that can be performed during training \cite{guo2019reweighted, qiu2019blockwise}. These successful results are corroborated by other efforts \cite{gordon-etal-2020-compressing} showing that low levels of pruning do not substantially affect pretrained representations. Additional successful efforts in model pruning directly target a downstream task \cite{sun-etal-2019-patient, Michel2019AreSH, McCarley2019PruningAB, cao-etal-2020-deformer}. Note that pruning does not always lead to speedups in practice as sparse operations may be hard to parallelize.

Knowledge distillation (KD) uses an overparameterized teacher model to rapidly train a smaller student model with minimal loss in performance \cite{Hinton2015DistillingTK} and has been used for translation \cite{kim-rush-2016-sequence}, computer vision \cite{howard2017mobilenets}, and adversarial examples \cite{Carlini2016TowardsET}. This has been applied to ELMo \cite{Li2019Efficient} and BERT \citep[\textit{inter alia}]{tang-etal-2019-natural, Sanh2019DistilBERTAD, sun-etal-2020-mobilebert}. KD can also be combined with adaptive inference, which dynamically adjusts model size \cite{liu2020fastbert}, or performed on submodules which are later substituted back into the full model \cite{xu2020bertoftheseus}.

Quantization with custom low-precision hardware is also a promising method for both reducing the size of models and compute time, albeit it does not reduce the number of parameters or FPOs \cite{Shen2019QBERTHB, Zafrir2019Q8BERTQ8}. This line of work is mostly orthogonal to other efforts specific to NLP. 

\subsection{Standardizing Comparison}
\label{sec:efficiency:standard}

There has yet to be a comprehensive and fair evaluation across all models. The closest, Table 1 in \citet{rogers2020primer}, compares 12 works in model compression. \textbf{However, almost no two papers are evaluated against the same BERT with the same set of tasks.} Many papers on attention sparsification do not evaluate on NLU benchmarks. We claim this is because finetuning is itself an expensive task, so it is not prioritized by authors: works on improving model efficiency have focused only on comparing to a BERT on a few tasks.

While it is easy for future research on pretraining to report model sizes and runtimes, it is harder for researchers in efficiency to report NLU benchmarks. We suggest extending versions of the leaderboards under different resource constraints so that researchers with access to less hardware could still contribute under the resource-constrained conditions. Some work has begun in this direction: the SustaiNLP 2020 Shared Task is focused on the energy footprint of inference for GLUE.\footnote{\url{https://sites.google.com/view/sustainlp2020/shared-task}}

\section{Area III: (Pretraining) Data}
\label{sec:data}

Unsurprisingly for our field, increasing the size of training data for an encoder contributes to increases in language understanding capabilities \cite{Yang19XLNet, raffel2019exploring, kaplan2020scaling}. At current data scales, some models converge before consuming the entire corpus. In this section, we identify a weakness when given \textit{less} data, advocate for better data cleaning, and raise technical and ethical issues with using web-scraped data. 

\subsection{Data Quantity}

There has not yet been observed a ceiling to the amount of data that can still be effectively used in training \cite{baevski-etal-2019-cloze, Liu19RoBERTa, Yang19XLNet, brown2020language}. \citet{raffel2019exploring} curate a 745GB subset of Common Crawl (CC),\footnote{\url{https://commoncrawl.org/} scrapes publicly accessible webpages each month.} which starkly contrasts with the 13GB used in BERT. For multilingual text encoding, \citet{wenzek-etal-2020-ccnet} curate 2.5TB of language-tagged CC. As CC continues to grow, there will be even larger datasets \cite{brown2020language}.

\citet{sun17revisiting} explore a similar question for computer vision, as years of progress iterated over 1M labeled images. By using 300M images, they improved performance on several tasks with a basic model. We echo their remarks that we should be cognizant of data sizes when drawing conclusions.

Is there a floor to the amount of data needed to achieve current levels of success on language understanding benchmarks? As we decrease the data size, LSTM-based models start to dominate in perplexity \cite{Yang19XLNet, Melis2020Mogrifier}, suggesting there are challenges with either scaling up LSTMs or scaling down Transformers. While probing contextualized models and representations is an important area of study (see \S\ref{sec:interpretability}), prior work focuses on pretrained models or models further pretrained on domain-specific data \cite{gururangan-etal-2020-dont}. We are not aware of any work which probes identical models trained with decreasingly less data. How much (and which) data is necessary for high performance on probing tasks?\footnote{\citet{conneau-etal-2020-unsupervised} claim we need a few hundred MiB of text data for BERT.}

\subsection{Data Quality}

While text encoders should be trained on language, large-scale datasets may contain web-scraped and uncurated content (like code). \citet{raffel2019exploring} ablate different types of data for text representations and find that naively increasing dataset size does not always improve performance, partially due to data quality. This realization is not new. Parallel data and alignment in machine translation \cite[\emph{inter alia}]{moore-lewis-2010-intelligent, duh-etal-2013-adaptation, xu-koehn-2017-zipporah, koehn-etal-2018-findings} and speech \cite{peddinti2016far} often use language models to filter out misaligned or poor data. \citet{sun17revisiting} use automatic data filtering in vision. These successes on other tasks suggest that improved automated methods of data cleaning would let future models consume more \textit{high-quality} data.

In addition to high quality, data uniqueness appears to be advantageous. \citet{raffel2019exploring} show that increasing the repetitions (number of epochs) of the pretraining corpus hurts performance. This is corroborated by \citet{Liu19RoBERTa}, who find that random, unique masks for MLM improve over repeated masks across epochs. These findings together suggest a preference to seeing more \textit{new} text. We suspect that representations of text spans appearing multiple times across the corpus are better shaped by observing them in unique contexts.

\citet{raffel2019exploring} find that differences in domain mismatch in pretraining data (web crawled vs. news or encyclopedic) result in strikingly different performance on certain challenge sets, and \citet{gururangan-etal-2020-dont} find that continuing pretraining on both domain and task specific data lead to gains in performance.

\subsection{Datasets and Evaluations}

With these larger and cleaner datasets, future research can better explore tradeoffs between size and quality, as well as strategies for scheduling data during training.

As we continue to scrape data off the web and publish challenge sets relying on other web data, we need to cautiously construct our training and evaluation sets. For example, the domains of many benchmarks (\citet[GLUE]{wang2018glue}, \citet[SQuAD]{rajpurkar-etal-2016-squad, rajpurkar-etal-2018-know}, \citet[SuperGLUE]{NIPS2019_8589}, \citet[LAMBADA]{paperno-etal-2016-lambada}, \citet[CNN/DM]{nallapati-etal-2016-abstractive}) now overlap with the data used to train language representations. Section 4 in \citet{brown2020language} more thoroughly discuss the effects of overlapping test data with pretraining data. \citet{gehman2020realtoxicityprompts} highlight the prevalance of toxic language in the common pretraining corpora and stress the important of pretraining data selection, especially for deployed models. We are not aware of a comprehensive study that explores the effect of leaving out targeted subsets of the pretraining data. We hope future models note the domains of pretraining and evaluation benchmarks, and for future language understanding benchmarks to focus on more diverse \textit{genres} in addition to diverse \textit{tasks}.

As we improve models by training on increasing sizes of crawled data, these models are also being picked up by NLP practitioners who deploy them in real-world software. These models learn biases found in their pretraining data \cite[\textit{inter alia}]{gonen-goldberg-2019-lipstick, may-etal-2019-measuring}. \textbf{It is critical to clearly state the source\footnote{How was the data generated, curated, and processed?} of the pretraining data and clarify appropriate uses of the released models.} For example, crawled data can contain incorrect facts about living people; while webpages can be edited or retracted, publicly released ``language'' model are frozen, which can raise privacy concerns \cite{10.1145/3336191.3371881}.

\section{Area IV: Interpretability}

\label{sec:interpretability}
While it is clear that the performance of text encoders surpass human baselines, it is less clear what knowledge is stored in these models; how do they make decisions? 
In their survey, \citet{rogers2020primer} find answers to the first question and also raise the second. Inspired by prior work \cite{lipton2016mythos, belinkov-glass-2019-analysis, Alishahi2019AnalyzingAI}, we organize here the major probing \textit{methods} that are applicable to all encoders in hopes that future work will use comparable techniques.

\subsection{Probing with Tasks}

One technique uses the learned model as initialization for a model trained on a \textit{probing task} consisting of a set of targeted natural language examples. The probing task's format is flexible as additional, (simple) diagnostic classifiers are trained on top of a typically frozen model \cite{ettinger-etal-2016-probing, Hupkes2018diagno, poliak-etal-2018-collecting-diverse, tenney2018what}. 
Task probing can also be applied to the embeddings at various layers to explore the knowledge captured at each layer \cite{tenney-etal-2019-bert,lin-etal-2019-open,liu-etal-2019-linguistic}. \citet{hewitt-liang-2019-designing} warn that expressive (nonlinear) diagnostic classifiers can learn more arbitrary information than constrained (linear) ones. This revelation, combined with the differences in probing task format and the need to train, leads us to be cautious in drawing  conclusions from these methods.

\subsection{Model Inspection}

Model inspection directly opens the metaphorical black box and studies the model weights without additional training.  For examples, the embeddings themselves can be analyzed as points in a vector space \cite{ethayarajh-2019-contextual}. Through visualization, attention heads have been matched to linguistic functions \cite{vig-2019-multiscale, clark-etal-2019-bert}. These works suggest inspection is a viable path to debugging specific examples. In the future, methods for analyzing and manipulating attention in machine translation \cite{lee-etal-2017-interactive, liu-etal-2018-visual, bau2018identifying, voita-etal-2019-bottom} can also be applied to text encoders.

Recently, interpreting attention as explanation has been questioned \cite{serrano-smith-2019-attention, jain-wallace-2019-attention, wiegreffe-pinter-2019-attention, clark-etal-2019-bert}. The ongoing discussion suggests that this method may still be insufficient for uncovering the rationale for predictions, which is critical for real-world applications.

\subsection{Input Manipulation\footnote{This is analogous to the ``few-shot`` and ``zero-shot'' analysis in \citet{brown2020language}.}}

Input manipulation draws conclusions by recasting the probing task format into the form of the pretraining task and observing the model's predictions. As discussed in \S\ref{sec:pretraining}, word prediction (cloze task) is a popular objective. This method has been used to investigate syntactic and semantic knowledge \cite{Goldberg2019AssessingBS, ettinger2019bert, kassner2019negated}. For a specific probing task, \citet{warstadt-etal-2019-investigating} show that cloze and diagnostic classifiers draw similar conclusions. As input manipulation is not affected by variables introduced by probing tasks and is as interpretable than inspection, we suggest more focus on this method: either by creating new datasets \cite{warstadt2019blimp} or recasting existing ones \cite{brown2020language} into this format. A disadvantage of this method (especially for smaller models) is the dependence on both the pattern used to elicit an answer from the model and, in the few-shot case where a couple examples are provided first, highly dependent on the examples \cite{schick2020just}.

\subsection{Future Directions in Model Analysis}
\label{sec:interpretability:future}

Most probing efforts have relied on diagnostic classifiers, yet these results are being questioned. Inspection of model weights has discovered what the models learn, but cannot explain their causal structure. We suggest researchers shift to the paradigm of input manipulation. By creating cloze tasks that assess linguistic knowledge, we can both observe decisions made by the model, which would imply (lack of) knowledge of a phenomenon. Furthermore, it will also enable us to directly interact with these models (by changing the input) without additional training, which currently introduces additional sources of uncertainty.

\citet{bender-koller-2020-climbing} also recommend a top-down view for model analysis that focuses on the end-goals for our field over hill-climbing individual datasets. While language models continue to outperform each other on these tasks, they argue these models do not learn \textit{meaning}.\footnote{A definition is given in \S 3 of \citet{bender-koller-2020-climbing}.} If not meaning, what are these models learning? 

We are overinvesting in BERT. While it is fruitful to understand the boundaries of its knowledge, we should look more across (simpler) models to see \textit{how} and \textit{why} specific knowledge is picked up as our models both become increasingly complex and perform better on a wide set of tasks. For example, how many parameters does a Transformer-based model need to outperform ELMo or even rule-based baselines?

\section{Area V: Multilinguality}
\label{sec:multilinguality}

The majority of research on text encoders has been in English.\footnote{Of the monolingual encoders in other languages, core research in modeling has only been performed so far for a few non-English languages \cite{Sun2019ERNIEER,sun2019ernie}.} Cross-lingual shared representations have been proposed as an efficient way to target multiple languages by using multilingual text for pretraining \cite[\textit{inter alia}]{mulcaire-etal-2019-polyglot, devlin-etal-2019-bert, lample2019cross, liu2020multilingual}.
For evaluation, researchers have devised multilingual benchmarks mirroring those for NLU in English \cite{conneau-etal-2018-xnli, liang2020xglue, hu2020xtreme}. Surprisingly, without any explicit cross-lingual signal, these models achieve strong zero-shot cross-lingual performance, outperforming prior cross-lingual word embedding-based methods \cite{wu-dredze-2019-beto, pires-etal-2019-multilingual}.

A natural follow-up question to ask is why these models learn cross-lingual representations. Some answers include the shared subword vocabulary \cite{pires-etal-2019-multilingual, wu-dredze-2019-beto}, shared Transformer layers \cite{conneau-etal-2020-emerging,artetxe-etal-2020-cross} across languages, and depth of the network \cite{K2020Cross-Lingual}. Studies have also found the geometry of representations of different languages in the multilingual encoders can be aligned with linear transformations \cite{schuster-etal-2019-cross,wang-etal-2019-cross,Wang*2020Cross-lingual,liu-etal-2019-investigating}, which has also been observed in independent monolingual encoders \cite{conneau-etal-2020-emerging}. These alignments can be further improved \cite{Cao2020Multilingual}.

\subsection{Evaluating Multilinguality}

All of the areas discussed in this paper are applicable to multilingual encoders. However, progress in training, architecture, datasets, and evaluations are occurring concurrently, making it difficult to draw conclusions. We need more comparisons between competitive multilingual and monolingual systems or datasets. To this end, \citet{wu-dredze-2020-languages} find that monolingual BERTs in low-resource languages are outperformed by multilingual BERT. Additionally, as zero-shot (or few-shot) cross-lingual transfer has inherently high variance \cite{keung2020evaluation}, \textbf{the variance of models should also be reported}.

We anticipate cross-lingual performance being a new dimension to
consider when evaluating text representations.  For
example, it will be exciting to discover how a small, highly-performant 
monolingual encoder contrasts against a multilingual variant;
e.g., what is the minimum number of parameters needed to support a new language? Or, how does model size relate to the phylogenetic diversity of languages supported?

\section{Discussion}
\label{sec:discussion}

\subsection{Limitations and Recommendations}
This survey, like others, is limited to only what has been shared publicly so far. The papers of many models described here highlight their best parts, where potential flaws are perhaps obscured within tables of numbers. Leaderboard submissions that do not achieve first place may never be published. Meanwhile, encoders are expensive to work with, yet they are a ubiquitous component in most modern NLP models. We strongly encourage more \textbf{publication and \textit{publicizing} of negative results} and limitations. In addition to their scientific benefits,\footnote{An EMNLP 2020 workshop is motivated by better science  (\url{https://insights-workshop.github.io/}).} publishing negative results in contextualized encoders can avoid significant externalities of rediscovering what doesn't work: time, money, and electricity. Furthermore, we ask leaderboard owners to \textbf{periodically publish surveys} of their received submissions.

The flourishing research in improving encoders is rivaled by research in interpreting them, mainly focused on discovering the boundary of what knowledge is captured by the models. For investigations that aim to sharpen the boundary, it is logical to build off of these prior results. However, we raise a concern that these encoders are all trained on similar data and have similar sizes. Future work in \textbf{probing should also look across different sizes and domains of training data}, as well as study the effect of model size. This can be further facilitated by model creators who release (data) ablated versions of their models.

We also raise a concern about reproducibility and accessibility of evaluation. Already, several papers focused on model compression do not report full GLUE results, possibly due to the expensive finetuning process for each of the nine datasets. Finetuning currently requires additional compute and infrastructure,\footnote{\citet{pruksachatkun-etal-2020-jiant} is a library that reduces some infrastructural overhead of finetuning.} and the specific methods used impact task performance. As long as finetuning is still an essential component of evaluating encoders, devising \textbf{cheap, accessible, and reproducible metrics for encoders is an open problem}. 

\citet{ribeiro-etal-2020-beyond} suggest a practical solution to both probing model errors and reproducible evaluations by creating tools that quickly generate test cases for linguistic capabilities and find bugs in models. This task-agnostic methodology may be extensible to both challenging tasks and probing specific linguistic phenomenon.

\subsection{Which *BERT should we use?}
Here, we discuss tradeoffs between metrics and synthesize the previous sections. We provide a series of questions to consider when working with encoders for research or application development. 

\paragraph{Task performance vs. efficiency} An increasingly popular line of recent work has investigated knowledge distillation, model compression, and sparsification of encoders (\S\ref{sec:efficiency:space}). These efforts have led to significantly smaller encoders that boast competitive performance, and under certain settings, non-contextual embeddings alone may be sufficient \cite{arora-etal-2020-contextual, wang-etal-2020-pretrain}. For downstream applications, ask: \textbf{Is the extra iota of performance worth the significant costs of compute?}

\paragraph{Leaderboards vs. real data} As a community, we are hill-climbing on curated benchmarks that aggregate dozens of tasks. Performance on these benchmarks does not necessarily reflect that of specific real-world tasks, like understanding social media posts about a pandemic \cite{mller2020covidtwitterbert}. Before picking the best encoder determined by average scores, ask: \textbf{Is this encoder the best for our specific task? Should we instead curate a large dataset and pretrain again?} \citet{gururangan-etal-2020-dont} suggest continued pretraining on in-domain data as a viable alternative to pretraining from scratch.

For real-world systems, practitioners should be especially conscious of the datasets on which these encoders are pretrained. \textbf{There is a tradeoff between task performance and possible harms contained within the pretraining data.} 

\paragraph{Monolingual vs. Multilingual} For some higher resource languages, there exist monolingual pretrained encoders. For tasks in those languages, those encoders are a good starting point. However, as we discussed in \S\ref{sec:multilinguality}, multilingual encoders can, surprisingly, perform competitively, yet these metrics are averaged over multiple languages and tasks. Again, we encourage looking at the relative \textbf{performance for a specific task and language}, and whether \textbf{monolingual encoders (or embeddings) may be more suitable.}

\paragraph{Ease-of-use vs. novelty} With a constant stream of new papers and models (without peer review) for innovating in each direction, we suggest using and building off \textbf{encoders that are well-documented with reproduced or reproducible results}. Given the pace of the field and large selection of models, unless aiming to reproduce prior work or improve underlying encoder technology, we recommend proceeding with caution when reimplementing ideas from scratch.

\section{Conclusions}
In this survey we categorize research in contextualized encoders and discuss some issues regarding its conclusions. We cover background on contextualized encoders, pretraining objectives, efficiency, data, approaches in model interpretability, and research in multilingual systems. As there is now a large selection of models to choose from, we discuss tradeoffs that emerge between models.  We hope this work provides some assistance to both those entering the NLP community and those already using contextualized encoders in looking beyond SOTA (and Twitter) to make more educated choices. 

\section*{Acknowledgments}

We especially thank the (meta-)reviewers for their insightful feedback and criticisms. In addition, we thank Sabrina Mielke, Nathaniel Weir, Huda Khayrallah, Mitchell Gordon, and Shuoyang Ding for discussing several drafts of this work. This work was supported in part by DARPA AIDA (FA8750-18-2-0015) and IARPA BETTER (\#2019-19051600005). The views and conclusions contained in this work are those of the authors and should not be interpreted as necessarily representing the official policies, either expressed or implied, or endorsements of DARPA, ODNI, IARPA, or the U.S. Government. The U.S. Government is authorized to reproduce and distribute reprints for governmental purposes notwithstanding any copyright annotation therein.

\bibliography{ref} 
\bibliographystyle{acl_natbib}

\end{document}